\documentclass[10pt,twocolumn,letterpaper]{article}

\usepackage{cvpr}              

\usepackage[dvipsnames]{xcolor}

\definecolor{cvprblue}{rgb}{0.21,0.49,0.74}
\usepackage[pagebackref,break links,colorlinks,citecolor=cvprblue]{hyperref}
\usepackage{multirow}
\usepackage{multicol}
\usepackage[accsupp]{axessibility}
\usepackage{varwidth}
\usepackage{algorithm}
\usepackage{algpseudocode}

\def\paperID{10313} 
\def\confName{CVPR}
\def\confYear{2024}

\title{Towards Generalizing to Unseen Domains with Few Labels}

\author{Chamuditha Jayanga Galappaththige$^{1}$,  Sanoojan Baliah$^{1}$,  Malitha Gunawardhana$^{1,2}$, \\ Muhammad Haris Khan$^{1}$\\
$^1$Mohamed Bin Zayed University of Artificial Intelligence, UAE \hfill  
$^2$University of Auckland, New Zealand\\
\tt\small{\{chamuditha.jayanga,sanoojan.baliah,malitha.gunawardhana,muhammad.haris\}@mbzuai.ac.ae}
}

\begin{document}
\maketitle
\def \thefootnote{*}\footnotetext{Equally contributing authors}

\begin{abstract}

We approach the challenge of addressing semi-supervised domain generalization (SSDG). Specifically, our aim is to obtain a model that learns domain-generalizable features by leveraging a limited subset of labelled data alongside a substantially larger pool of unlabeled data. Existing domain generalization (DG) methods which are unable to exploit unlabeled data perform poorly compared to semi-supervised learning (SSL) methods under SSDG setting. Nevertheless, SSL methods have considerable room for performance improvement when compared to fully-supervised DG training. To tackle this underexplored, yet highly practical problem of SSDG, we make the following core contributions. First, we propose a feature-based conformity technique that matches the posterior distributions from the feature space with the pseudo-label from the model's output space. Second, we develop a semantics alignment loss to learn semantically-compatible representations by regularizing the semantic structure in the feature space. Our method is plug-and-play and can be readily integrated with different SSL-based SSDG baselines without introducing any additional parameters. Extensive experimental results across five challenging DG benchmarks with four strong SSL baselines suggest that our method provides consistent and notable gains in two different SSDG settings. Our code is available at \href{https://github.com/Chumsy0725/FBC-SA/}{FBC-SA}.

\end{abstract}

\begin{figure}
     \centering
     \includegraphics[width=1\linewidth]{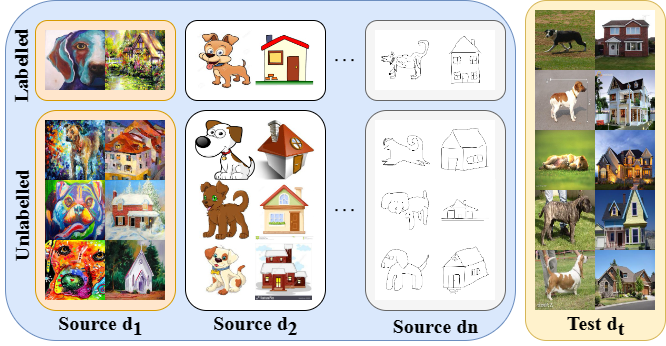}
     \caption{Visual illustration of Semi-supervised Domain Generalization (SSDG) setting.}
     \label{fig:illustration}
      \vspace*{-\baselineskip}
 \end{figure}
\vspace{-0.5em}    
\section{Introduction}
\label{section:Introduction}



Top-performing visual object recognition models \cite{simonyan2014very, he2016deep, dosovitskiy2020image} tend to sacrifice performance when there is a difference between the training and testing data distributions, also termed as the domain shift problem \cite{zhang2013domain, zhang2022latent}.
To handle the domain shift problem, the research line of domain generalization (DG), amongst others, has been explored with greater interest. As such, DG places fewer assumptions and so it is potentially more widely applicable than the other alternatives e.g., unsupervised domain adaptation (UDA) \cite{zhang2021survey}. 
%
In DG, the goal is to learn a generalizable model, relying on data from multiple source domains for training, that is capable of performing well on data from an unseen target domain \cite{blanchard2011generalizing, muandet2013domain, li2017deeper}. In the recent past, several promising DG methods have been proposed that leverage different feature-level constraints \cite{muandet2013domain}, adapt meta-learning frameworks \cite{balaji2018metareg, li2019episodic}, leverage proxy tasks \cite{carlucci2019domain, wang2020learning}, or develop data augmentation mechanisms \cite{zhou2020learning, khan2021mode}. Note that almost all existing DG methods operate under fully supervised settings i.e. the data from all source domains are completely labeled.

In many real-world use-cases, for instance, in healthcare, it is often hard, if not impossible, to acquire a sufficiently large set of labeled data from all source domains \cite{yang2022survey}. Usually, only a small subset of data is labelled, and the remaining large fraction of data is unlabelled. Therefore, besides, aiming to achieve cross-domain generalization, a model should be able to rely on limited labels \cite{yuan2022label}. 
The topic of semi-supervised learning (SSL) is quite relevant in this case \cite{lee2013pseudo, tarvainen2017mean, sohn2020fixmatch, berthelot2019mixmatch}, which aims to leverage abundantly available unlabeled data with a small fraction of labeled data to achieve learning.

\begin{figure}
     \centering
     \includegraphics[width=1\linewidth]{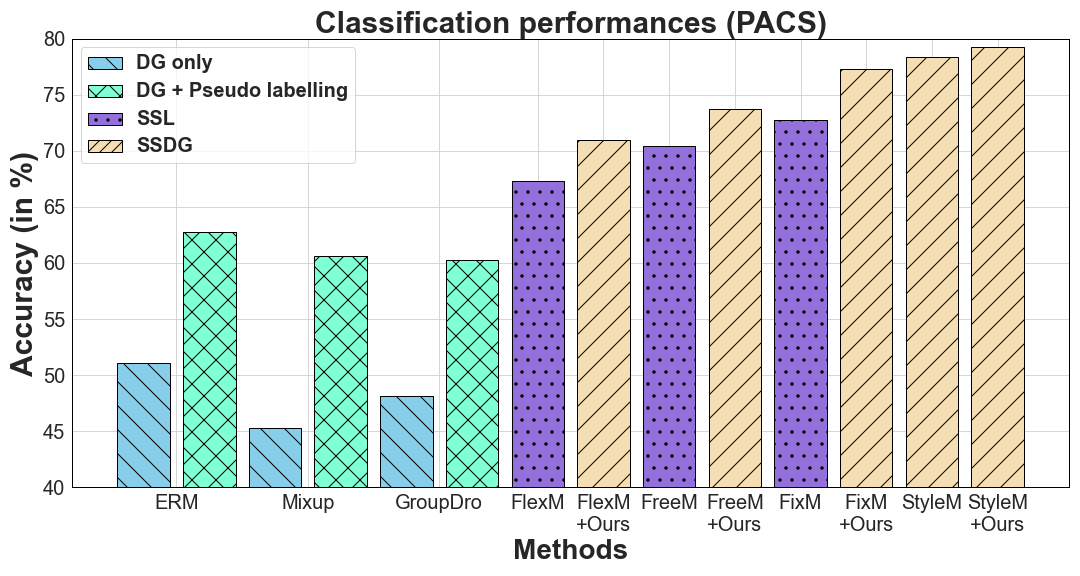}
     \caption{Recognition performance comparison between different DG, DG combined with pseudo-labeling, SSL methods and ours in SSDG settings. Here, GD - GroupDro, FlexM - FlexMatch, FreeM - FreeMatch, FixM - FixMatch and StyleM - StyleMatch.}
     \label{fig:pacs-5label}
     \vspace*{-\baselineskip}
 \end{figure}

In our work, we aim to study the problem of semi-supervised domain generalization (SSDG), which unifies domain generalization and data efficiency under the same framework \cite{zhou2023semi}. SSDG is similar to DG in terms of core objective i.e. learning a generalizable model by leveraging multiple source domains. However, DG assumes that the data from all source domains are fully-labelled. On the contrary, SSDG operates under SSL settings where only a handful of labeled data is available while a relatively large chunk of data is unlabeled \cite{zhou2023semi}. Fig.~\ref{fig:illustration} draws a visual illustration of the SSDG setting. It is shown that the DG methods tend to perform poorly under the limited labels setting of SSDG as they are not developed to exploit the unlabeled data. On the other hand, SSL methods, in particular, \cite{sohn2020fixmatch}, perform relatively better than the DG methods \cite{zhou2023semi}, but still their obtained performance is considerably lower than the fully supervised training. Fig.~\ref{fig:pacs-5label} shows the comparison of results in SSDG problem setting among the following methods: DG, DG combined with pseudo-labeling (following \cite{sohn2020fixmatch}) and SSL. Note that, the naive combination of DG and SSL methods also performs poorly \cite{zhou2023semi}. 

We propose a new approach for tackling semi-supervised domain generalization (SSDG) by observing the key limitations in best-performing SSL-based baselines for SSDG. A dominant challenge is how to achieve accurate pseudo-labels (PLs) when the unlabeled data exhibit various domain shifts. It is further exacerbated by the scarcity of labeled data, which possibly increases the chances of model overfitting. Towards addressing them, we resort to the feature space of the model and propose a feature-based conformity (FBC) module and a semantics alignment (SA) loss. 
The feature-based conformity module aims at aligning the posteriors from the feature space with the pseudo-labels from the model's output space via two different constraints. To learn semantically harmonious features under unlabelled data from multiple heterogeneous sources, we present a semantics alignment loss that attempts to regularize the semantic structure in the feature space by domain-aware similarity-guided cohesion and repulsion of examples. In summary, we make the following key contributions:

\begin{itemize}[topsep=-3pt, noitemsep]

\item We study the relatively unexplored yet highly practical problem of semi-supervised domain generalization (SSDG) and propose a new approach, consisting of feature-based conformity and semantics alignment loss, for addressing the important challenges in SSDG. 

\item Our approach is plug-and-play and as such it can be seamlessly applied to different SSL-based SSDG baselines without adding any learnable parameters. We show the adaptability and effectiveness of our method with four strong baselines.

\item We perform extensive experiments on five different DG datasets: PACS \cite{li2017deeper}, OfficeHome \cite{venkateswara2017deep}, DigitsDG \cite{zhou2020deep}, TerraIncognita \cite{beery2018recognition} and VLCS \cite{torralba2011unbiased} with four strong baselines:  FixMatch \cite{sohn2020fixmatch}, FlexMatch \cite{zhang2021flexmatch} and  FreeMatch \cite{wang2023freematch}  StyleMatch\cite{zhou2023semi}. Our approach delivers consistent and visible gains across all datasets with four baselines in two variants of SSDG settings.

\end{itemize}

\begin{figure*}
\centering
\begin{subfigure}{0.49\linewidth}
    \centering
    \includegraphics[width=1\linewidth]{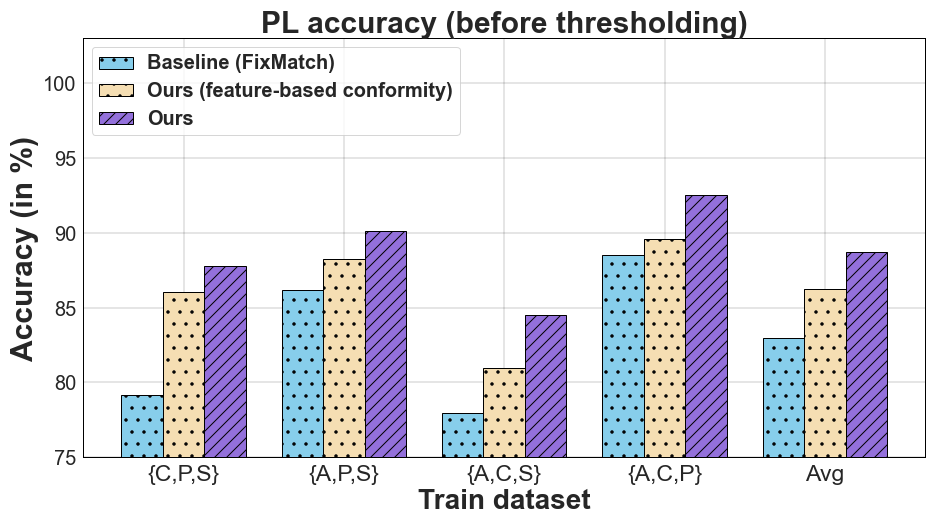}  
    \label{SUBFIGURE LABEL 2}
\end{subfigure}
\begin{subfigure}{0.49\linewidth}
    \centering
    \includegraphics[width=1\linewidth]{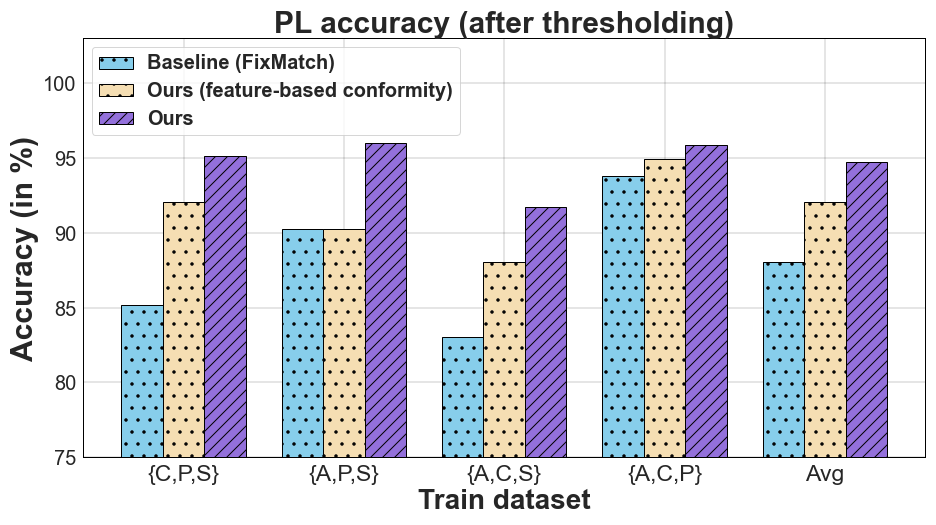}  
    \label{SUBFIGURE LABEL 1}
\end{subfigure}

 \vspace*{-\baselineskip}
\caption{PL accuracy in the training samples without thresholding (left) and for selected PL after thresholding (right) for the baseline (FixMatch \cite{sohn2020fixmatch}), ours with only feature-based conformity, and ours for PACS dataset in 5 labels per class setting. Here A, C, P, and S denote Art-painting, Cartoon, Photos, and Sketch domains, respectively.}

\vspace*{-\baselineskip}
\label{fig:PL_accuracy}
\end{figure*}



\section{Related Work}
\label{section:Related Work}

\noindent\textbf{Domain generalization:}
Several studies have been conducted towards improving domain generalization (DG) performance. Most methods aim to learn domain-invariant features from available source domain data \cite{muandet2013domain, ghifary2015domain}. Empirical risk minimization (ERM) can be regarded as the earliest attempt that aims to reduce the sum of errors across data aggregated from multiple source domains \cite{vapnik1999nature}. Following DG works, utilized maximum mean discrepancy (MMD) constraint \cite{muandet2013domain}, developed multi-task autoencoder \cite{ghifary2015domain}, achieved adversarial feature learning with MMD\cite{li2018domain}, learned invariant predictors \cite{arjovsky2019invariant}, introduced low-rank regularization \cite{li2017deeper,xu2014exploiting} in pursuit of extracting domain-invariant features. Another line of work adapted the meta-learning framework to simulate domain drifts during training \cite{li2019episodic}. Furthermore, some work leveraged proxy tasks \cite{carlucci2019domain, wang2020learning} to promote domain-generalizable features. Some used domain-specific masks \cite{chattopadhyay2020learning} and domain-specific normalizations \cite{seo2019learning} to strike the balance between domain-specific and domain-invariant features. Inspired by the contrastive learning paradigm, some DG works adapted self-supervision and different variants of ranking losses \cite{motiian2017unified,dou2019domain,Kim_2021_ICCV}. \cite{cha2021swad} proposed stochastic weight averaging in a dense manner to achieve flatter minima for DG. Many DG approaches proposed new techniques to synthesize examples from novel domains to increase the diversity in source domains. \cite{shankar2018generalizing} devised crossgrad training \cite{shankar2018generalizing}, \cite{volpi2018generalizing} imposed a wassertein constraint in semantic space, \cite{zhou2020learning} adapted a CNN-based generator. \cite{khan2021mode} exploited class-conditional covariance to augment novel source domain features. \cite{yan2020improve} used mixup for creating new images by mixing source images.

However, most of the existing DG works expect fully labelled data from all source domains. Their performance degrades substantially upon reducing the amount of labeled data (see Fig.~\ref{fig:pacs-5label}), which limits their applicability to several important application domains, including medical imaging and autonomous vehicles, where sufficiently labelled data is scarce. To this end, in this paper, we study the relatively underexplored problem of semi-supervised domain generalization (SSDG), which unifies the domain generalization and data efficiency under a common framework, and propose a new SSDG approach based on feature-based conformity and semantics alignment constraint.

\begin{figure*} 
    \centering
    \includegraphics[width=\textwidth]{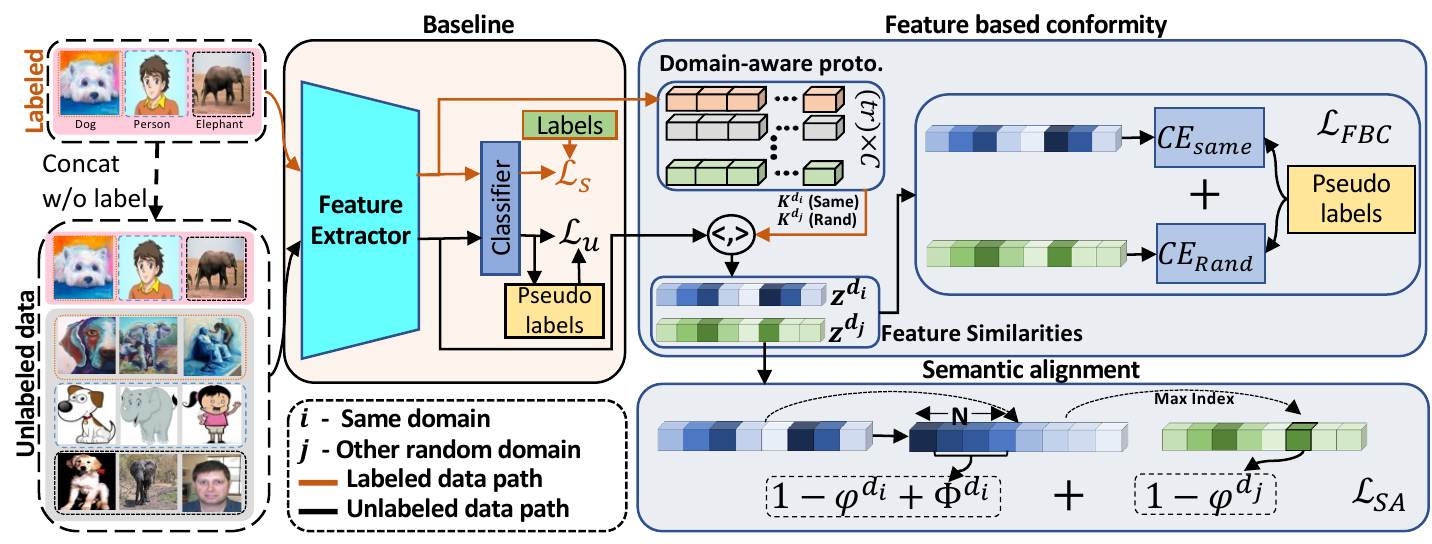}
    \caption{Overall architecture of our method. Fundamentally, it is a semi-supervised baseline (e.g., FixMatch \cite{sohn2020fixmatch}) with a feature extractor and a classifier that involves pseudo-labelling and prediction consistency mechanisms. To tackle semi-supervised domain generalization, we first propose a \emph{feature-based conformity} module (sec.~\ref{subsec:Feature_based_conformity}) that aligns the posterior from feature space with the pseudo-label from output space. We then develop a \emph{semantics alignment loss} (sec.~\ref{subsec: sematic_alignment_constraint}) to regularize the semantic layout of feature space and further improve the effectiveness of feature-based conformity.}
    \label{fig:method}
     \vspace*{-\baselineskip}
\end{figure*}

\noindent\textbf{Semi-Supervised Learning:}
The research direction of Semi-Supervised Learning (SSL) has seen numerous methods developed in the literature, with entropy minimization \cite{grandvalet2004semi} consistency learning \cite{miyato2018virtual,tarvainen2017mean, xie2020unsupervised} and pseudo-labelling \cite{lee2013pseudo,sohn2020fixmatch} being the most prominent approaches. Consistency learning involves making predictions of a model on two different views of the same input similar to each other \cite{zhou2003learning} by imposing a consistency loss on penultimate features \cite{abuduweili2021adaptive} or output probabilities \cite{sohn2020fixmatch}. Recently, \cite{tarvainen2017mean} found that using a model's exponential moving average to generate the target for consistency learning improves performances.
On the other hand, pseudo-labeling \cite{lee2013pseudo} generates either soft or hard pseudo-labels for unlabeled data using a pre-trained model \cite{xie2020unsupervised} or the model under training \cite{sohn2020fixmatch}. 
Further,  \cite{xie2020unsupervised,sohn2020fixmatch,xie2020self,berthelot2019mixmatch} demonstrated that inducing strong noise such as the strong augmentation or dropout to the student model can significantly boost performance. \cite{chen2023softmatch,wang2023freematch,zhang2021flexmatch} focus on improving pseudo-labelling building upon \cite{sohn2020fixmatch}. To address distribution shifts between labeled and unlabeled data caused by sampling bias, some studies \cite{abuduweili2021adaptive,wang2019high} have adopted ideas from domain adaptation \cite{hoffman2018cycada} to minimize the feature distance. 

While semi-supervised domain generalization (SSDG) and SSL both deal with unlabelled data, SSDG poses a greater challenge as the data is collected from heterogeneous sources with potentially different underlying data distributions. SSDG is a relatively underexplored problem and so very little research has been done. \cite{yeGraph} proposed active exploration, which queries labels of examples with higher ranks in class uncertainty, domain representativeness and information diversity, and combines inter and intra-domain knowledge with mixup \cite{zhang2017mixup}. \cite{yuan2022label} proposed a method that relies on a generated similarity graph and a graph Laplacian regularizer. \cite{wang2023better} proposed a joint domain-aware label and dual classifier framework for learning a domain-generalizable model when only one source domain is fully labelled while the others are completely unlabelled. Recently, Stylematch \cite{zhou2023semi} extends \cite{sohn2020fixmatch} for SSDG with stochastic modeling to reduce overfitting and multi-view consistency learning for generalizing across domains. Multi-view consistency operates within the confines of the input pixel space and assumes some style variance between the data distributions. Such an approach struggles or is unable to improve performance when presented with different types of distributions such as background shifts or corruption shifts as shown in our experiments (sec.~\ref{section:Experiments}). In this work, we propose a new SSDG approach, for partially-labelled source domains, that explores the feature space to develop a feature-based conformity mechanism and a semantics alignment constraint. Feature space provides more flexibility compared to input pixel space for imposing various consistencies. This is beneficial towards effectively tackling SSDG challenges as demonstrated in our results.

\section{Methodology}
\label{section:Methodology}

\subsection{Preliminaries}

\noindent\textbf{Problem Settings:} 
We adapt some notation from \cite{Gulrajani2021InSO}. Similar to multi-source DG settings, 
let us characterize each domain $d$ by $d=\{(\mathbf{x}_i^{d},y_i^d)\}_{i=1}^{n}$, where $\mathbf{x}^d_i \in \mathbb{R}^{C \times H \times W}$ is an input image, $y^d_i$ is the corresponding label, and comprised of $n$\footnote{The value of $n$ can be different for each domain.} 
independent and identically distributed (i.i.d) 
examples drawn from a joint probability distribution $\mathcal{P}(\mathcal{X}^d,\mathcal{Y}^d)$ for all possible training (source) domains $ d \in \{d_1,..., d_{tr}\}$.  
$\mathcal{X}^{d}$ is an input space over which the domain $d$ is defined and $\mathcal{Y}^{d}$ is the corresponding label space. Here, we consider the distribution shift in $\mathcal{P}(\mathcal{X}^d)$ while $\mathcal{P}(\mathcal{Y}^d)$ shares the same label space $\mathcal{Y}$ where $y \in \mathcal{Y} = \{1,2,..,C\}$ is an associated class label.

In the SSDG setting, the number of labeled examples is constrained i.e., each source domain $d$ has a labeled part $d^L=\{(\mathbf{x}_i^{d},y_i^{d})\}$ and an unlabeled part $d^U=\{(\mathbf{u}_i^{d})\}$. Further, the number of examples in the unlabeled part is much higher 
than in the labeled part, i.e., $\lvert d^U \rvert \gg \lvert d^L \rvert $. Our goal is to learn a domain-generalizable model $\mathcal{F}$ using the training (source) domains $ d \in \{d_1,..., d_{tr}\}$  to accurately predict on an out-of-distribution data, whose examples are drawn from $\mathcal{P}(\mathcal{X}^{d_{te}},\mathcal{Y}^{d_{te}})$, where $d_{te}$ 
represents the target domain. For our methodology, we decompose the model as $\mathcal{F}=w\circ f$,  $f: \mathbf{x} \rightarrow h$ is a feature encoder and $w:h \rightarrow y$ is a classifier. $\mathcal{F}$ maps input images to the target label space. 

\noindent\textbf{SSDG pipeline:} 
We overview FixMatch \cite{sohn2020fixmatch} which emerged as a top-performing SSL-based SSDG baseline in our empirical investigations (Fig.~\ref{fig:pacs-5label}). So we chose it as an example baseline to explain our method. However, as shown in the experimental results (sec.~\ref{section:Experiments}), our method is model-agnostic and can be applied to several SSL-based SSDG baselines. FixMatch, which was originally proposed for SSL, combines two prior SSL techniques: pseudo-labeling, and consistency regularization. 
Pseudo-labeling uses a model to generate artificial labels for unlabeled data which are obtained from the $\arg \max$ of the model's prediction probability. It retains only those artificial labels whose largest class probability falls above a predefined threshold. On the other hand, consistency regularization \cite{bachman2014learning} leverages unlabeled data by enforcing that the predictions for perturbed views of the same image should be similar.

FixMatch\cite{sohn2020fixmatch} applies a weak augmentation and a strong augmentation to all images in a minibatch. The overall loss function consists of a supervised loss $\mathcal{L}_s$ and an unsupervised loss $\mathcal{L}_u$. $\mathcal{L}_s$ is a standard cross entropy (CE) loss applied on the weakly augmented labeled images. For unlabeled images, FixMatch first computes a pseudo-label corresponding to a weakly augmented version of the image and then uses this to enforce the cross-entropy loss against the model's output for a strongly augmented version of the same image, denoted as $\mathcal{L}_u$. This introduces a form of consistency regularization for the model. Overall the FixMatch loss is formulated as: $\mathcal{L}=\mathcal{L}_s+\mathcal{L}_u$.

\noindent\textbf{Discussion:} Although the SSL-based SSDG baselines show relatively better performance in SSDG settings, there is considerable room for further improvement across several DG benchmarks (Fig~\ref{fig:pacs-5label}). An important challenge, faced by them, is the selection of accurate pseudo-labels in the presence of multiple domain shifts (see Fig.~\ref{fig:PL_accuracy}). 
This is further aggravated by the scarcity of limited labels, which increases the chances of model overfitting. 
To tackle the SSDG problem, we leverage the feature space and propose to enforce the prediction consistency between the two quantities: feature space posteriors which are derived from the same and different domains, and the pseudo-label from the model's output space (sec.~\ref{subsec:Feature_based_conformity}). 
It implicitly facilitates the learning of more accurate pseudo-labels (see Fig.~\ref{fig:PL_accuracy}) by penalizing those examples whose prediction from the feature space does not align with the corresponding pseudo-label.
Towards further improving the model's discriminative ability under the SSDG setting, we develop a semantics alignment loss (sec.~\ref{subsec: sematic_alignment_constraint}) that attempts to regularize the semantic layout in the feature space by domain-aware similarity-guided cohesion and repulsion of training examples. Fig.~\ref{fig:method} visualizes the overall architecture of our method. In our method, described next, we ignore the labels of labeled set $d^L$  and merge it with the unlabeled set $d^U$ and so collectively we treat all the images as unlabeled afterward following \cite{zhou2023semi} unless stated otherwise. 

\begin{table*}[!htp]
\centering

\setlength{\tabcolsep}{15pt}
\scalebox{0.8}{
\begin{tabular}{lccccc}
\toprule
Model               & PACS  & OH & VLCS & DigitsDG & TerraInc.  \\
\midrule
ERM                 & $59.8\pm2.5$  & $56.7\pm0.8$ & $68.0\pm0.5$   & $29.1\pm2.9$   & $23.5\pm1.4$     \\ \midrule
EntMin              & $64.2\pm2.2$  & $57.0\pm0.8$ & $66.2\pm0.3$   & $39.3\pm2.8$  & $26.6\pm2.6$        \\
MeanTeacher         & $61.5\pm1.4$  & $55.9\pm0.5$ & $66.2\pm0.4$   & $38.8\pm2.9$  & $25.0\pm2.8$         \\
FlexMatch           & $72,7\pm1.2$  & $53.7\pm0.7$ & $56.2\pm2.1$   & $68.9\pm1.2$  & $26.4\pm1.8$        \\
FreeMatch           & $74.0\pm2.7$  & $56.2\pm0.2$ & $61.6\pm1.3$   & $67.5\pm2.4$  & $30.1\pm1.2$        \\
FixMatch            & $76.6\pm1.2$  & $57.8\pm0.3$ & $70.0\pm2.1$   & $66.4\pm1.4$  & $30.5\pm2.2$         \\
StyleMatch          & $79.4\pm0.9$  & $59.7\pm0.2$ & $73.5\pm0.6$   & $65.9\pm1.9$  & $29.9\pm2.8$          \\  \midrule
FlexMatch + Ours    & $75.3\pm1.2$  & $55.8\pm0.4$ & $58.7\pm1.0$   & $\textbf{73.1}\pm\textbf{1.1}$  & $30.9\pm1.0$          \\
FreeMatch + Ours    & $77.3\pm1.7$  & $58.0\pm0.4$ & $62.6\pm1.3$   & $72.2\pm0.4$  & $32.4\pm2.9$       \\
FixMatch + Ours     & $78.2\pm1.2$  & $59.0\pm0.4$ & $72.2\pm1.0$   & $70.4\pm1.4$  & $\textbf{34.7}\pm\textbf{1.9}$       \\
StyleMatch + Ours   & $\textbf{80.5}\pm\textbf{1.1}$  & $\textbf{60.3}\pm\textbf{0.6}$ & $\textbf{74.2}\pm\textbf{0.5}$   & $67.7\pm1.7$  & $32.5\pm1.8$  \\
\bottomrule
\end{tabular}}
\caption{SSDG accuracy (\%) with 10 labels per class. (Average over 5 independent seeds is reported.) }
\label{tab:10_labels}
\vspace{-\baselineskip}
\end{table*}

\subsection{Feature-based Conformity} 
\label{subsec:Feature_based_conformity}
We believe that, in the presence of multiple source domains manifesting various domain shifts, for an unlabeled example belonging to an arbitrary domain, the posterior from the same and different source domains in feature space, should align with its pseudo-label produced from the model's output space. In this work, we coin this as feature-based conformity (FBC) that implicitly facilitates the model towards generating more accurate pseudo-labels. To achieve feature-based conformity, for an unlabeled image feature, the following key steps are designed (Fig.~\ref{fig:method}). First, we build domain-aware class prototypes in the feature space with the features of labeled images. Second, to compute similarities for a given unlabeled image feature, we choose prototypes having the same domain label (referred as same-domain class prototypes) and also select another set of prototypes having a different domain label chosen randomly (referred as different-domain class prototypes). This yields same-domain and different-domain similarities. Third, we convert the same-domain and different-domain similarities to posterior probabilities, and finally, these two probabilities are aligned with the pseudo-labels. 

\noindent\textbf{Pseudo label generation:} For a given image $\mathbf{u}$ a pseudo label $\tilde{y}$ is generated by $\tilde{y}=\arg \max(\sigma \left( \mathcal{F} (\alpha(\mathbf{u})\right)))$ if $\max(\sigma \left( \mathcal{F} (\alpha(\mathbf{u})\right))) \geq \tau $, where $\sigma$ is a softmax function, $\alpha(.)$ represents the weak augmentation operation \cite{sohn2020fixmatch} and $\tau$ is a threshold to retain the most confident predictions as described in \cite{sohn2020fixmatch}.

\noindent\textbf{Domain-aware class prototypes:}
 We want to build domain-aware class prototypes which take into account the domain label $d$ and the class $c$. Specifically, we obtain the class prototypes by averaging the image features of labeled raw examples without any augmentation, from the penultimate layer, corresponding to the class $c$ and the domain $d$ in consideration. Thus, the domain-aware class prototype $\mathbf{K}_c^d$ for class $c$ in domain $d$ is defined as:
\begin{equation}
    \mathbf{K}_c^d = \frac{1}{\lvert \mathcal{S}_c^d  \rvert}\sum_{i=1}^{\lvert \mathcal{S}_c^d  \rvert} \mathcal{S}_c^d[i] \quad \text{where}  \, \,
    \mathcal{S}_c^d = \{f(\mathbf{x}_i^d)|({y}_i=c)\}_{i=1}^{\lvert d^L \rvert}.
\end{equation}
Here $\mathbf{x}_i^d$ is a labeled image from domain $d$ with class label $c$. Our domain-aware class prototypes are dynamically updated at the end of each epoch.

\noindent\textbf{Feature similarity and alignment:} Once we obtain the domain-aware class prototypes, we leverage them to obtain the same-domain and different-domain posterior probabilities in the feature space for the unlabeled images. 
To be more specific, for an image $\mathbf{u}^{d_i}$, we obtain similarities with the same-domain class prototypes and randomly chosen different-domain class prototypes. We compute the similarity with same-domain class prototype as $ \langle f(\alpha(\mathbf{u}^{d_i})), \mathbf{K}^{d_i}_{c} \rangle $, where $\langle,\rangle$ symbolizes the cosine similarity and $\alpha$ is the weak-augmentation function \cite{sohn2020fixmatch}. After computing the similarities with $C$ same-domain class prototypes, we get a vector of same-domain similarities, denoted as $\mathbf{z}^{d_i} \in \mathbb{R}^C$. We then take the softmax of $\mathbf{z}^{d_i}$ to obtain the same-domain posterior probability $p^{d_i}(\mathbf{K}^{d_i}_c|f(\alpha(\mathbf{u}^{d_i}))$.
Like-wise for obtaining similarity with the randomly chosen different-domain class prototype, we choose a random domain $d_j (\neq d_i)$ and compute $\langle f(\alpha(\mathbf{u}^{d_i})), \mathbf{K}^{d_j}_c \rangle $. After computing the similarities with $C$ different-domain class prototypes, we get a vector of different-domain similarities, denoted as $\mathbf{z}^{d_j} \in \mathbb{R}^C$. We take the softmax of $\mathbf{z}^{d_j}$ to obtain the different-domain posterior probability $p^{d_j}(\mathbf{K}^{d_j}_c|f(\alpha(\mathbf{u}^{d_i}))$.
Now, we leverage these same-domain probabilities  $\mathbf{p}^{d_i}$ and the different-domain probabilities $\mathbf{p}^{d_j}$ from feature space and propose to align with the pseudo-label $\tilde{y}$ using two cross-entropy losses:
\vspace{-0.4em}
\begin{equation}
    \mathcal{L}_{FBC}= -\underbrace{\sum_{i=1}^C \tilde{y}_i\mathrm{log}(p_{i}^{d_i}) }_{\text{Same domain}} -\underbrace{\sum_{i=1}^C\tilde{y}_i\mathrm{log}(p_{i}^{d_j})}_{\text{Random different domain}}  \label{eq:feature_view_constraint}  
\end{equation}

The feature-based conformity serves as a regularizer by aligning the \emph{the same-domain and different-domain probabilities} from the feature space with the corresponding pseudo-label.

\subsection{Semantics Alignment Constraint} 
\label{subsec: sematic_alignment_constraint}

\begin{figure*}
    \centering
    \includegraphics[width=\textwidth]{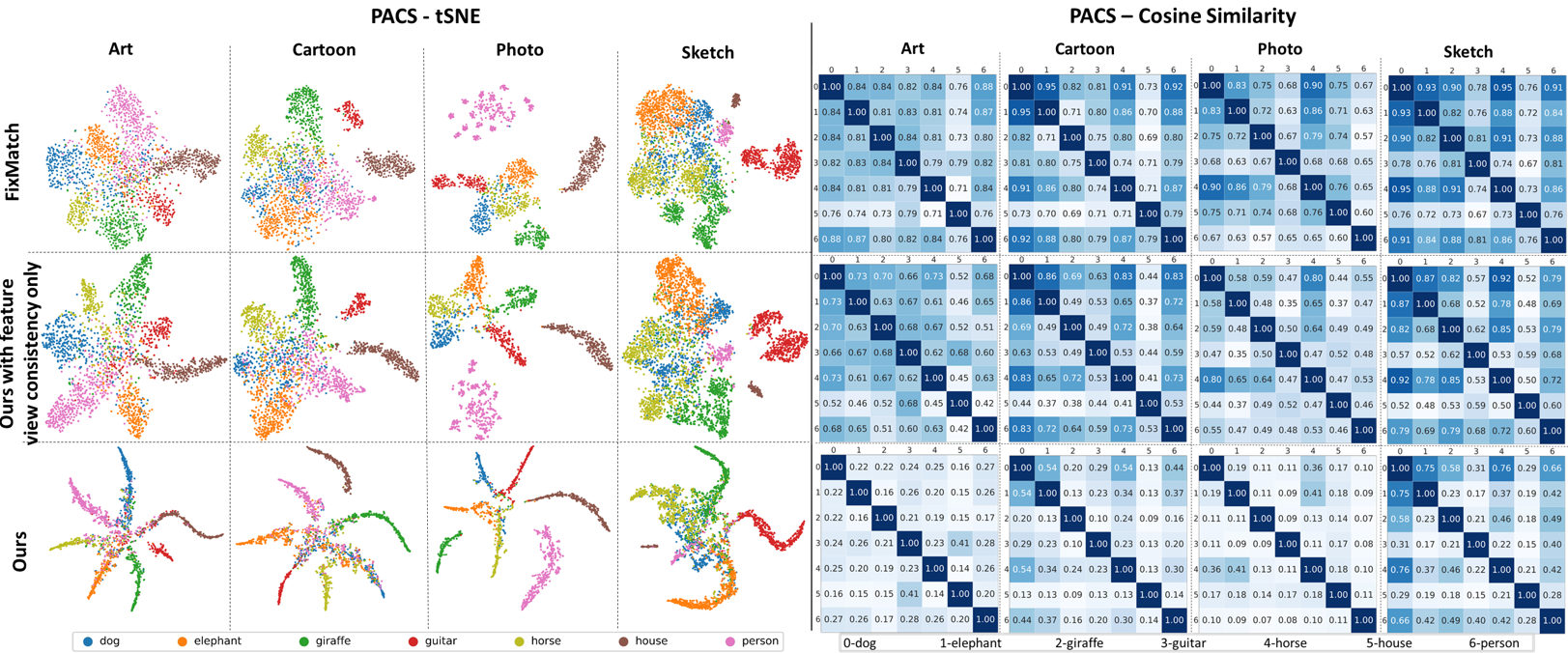}
    \caption{We visualize the feature space using tSNE (left), and the cosine similarity between the means of class-wise features (right) for PACS test domains of baseline (first row), ours with feature-based conformity only (second row), and ours with both feature-based conformity and semantic alignment loss (third row).} 
    
    \label{fig:tsne}
\end{figure*}

We intend to regularize the semantic structure in the feature space to capture semantically harmonious features and therefore improve the feature discriminativeness under domain shifts (Fig.~\ref{fig:tsne} left). To this end, we formulate a semantics alignment (SA) loss that achieves domain-aware similarity-guided cohesion and repulsion of training examples (Fig.~\ref{fig:method}). For an input feature, it attempts to maximize the similarity to the assigned prototype class while minimising the similarity to the hard non-assigned prototypes in the same domain. Hard non-assigned prototypes refer to the class prototypes which are nearest neighbors to the assigned prototype. Also, we maximize the similarity with a prototype of randomly chosen different domain, whereby this prototype is having the same class label as the assigned prototype in the same domain.
Formally, let $\phi^{d_i} = \max(\mathbf{z}^{d_i})$ and $ \phi^{d_j} = \mathbf{z}^{d_j}[\arg \max(\mathbf{z}^{d_i})]$ be the similarity corresponding to the assigned prototype in the same domain and the similarity indexed (with the assigned prototype) from different domain for an input feature corresponding to $\mathbf{u}^{d_i}$.
We also include the similarity to hard non-assigned prototypes in the same domain by first sorting same domain similarities  $\mathbf{z}^{d_i}$ (in descending order) to get $ \mathbf{v}^{d_i} = \mathrm{Sort}(\mathbf{z}^{d_i})$ and then selecting and averaging the top-N after excluding the highest similarity as: $\Phi^{d_i} = \frac{1}{N-1}\sum_{n=2}^{N}\mathbf{v}^{d_i}[n]$. With these quantities, our semantics alignment loss is formulated as:
\begin{equation}
\label{eq:SA}
    \mathcal{L}_{SA}=\underbrace{(1 - \phi^{d_i} + \Phi^{d_i})}_{\text{Same domain}} \quad \, + \underbrace{(1-\phi^{d_j})}_{\text{Random different domain}}
\end{equation}

To reduce the loss, the $\phi^{d_i}$ and $\phi^{d_j}$  terms will be maximized while the $\Phi^{d_i}$  term will be minimized. The same-domain component of the loss tries to align the feature to the assigned prototype and also tries to repel the hard non-assigned prototypes in the same domain. The different-domain component of the loss attempts to align the feature to the assigned prototype in different-domain. 

Our overall loss has four different loss terms: the supervised and unsupervised losses from the baseline, and the feature-based conformity loss and semantic alignment loss from our method, $ \mathcal{L}= \mathcal{L}_s +\mathcal{L}_u + \mathcal{L}_{FBC}+ \mathcal{L}_{SA}$. We provide the pseudo-code of our methodology in Algorithm.~\ref{alg:pseudo}.

\begin{algorithm}[!htp]
\caption{Pseudo-code}
\label{alg:pseudo}
\small
\begin{algorithmic}[1]
\State \textbf{Input:} Labeled batch $d^L=\{(\mathbf{x}_b^{d},y_b^{d}): b \in (1 , \cdots , B)\}$ and Unlabeled batch $d^U=\{(\mathbf{u}_b^{d}): b \in (1 , \cdots , B)\}$ $\forall d \in {d_1,d_2, \cdots d_{tr}}$, Confidence threshold $\tau$, E is total epochs, B is the Number of batches in an epoch, Model: $\mathcal{F}= w \circ f$ 
\For{epoch=1 to E}
    \State \parbox[t]{210pt}{\# Create domain-aware prototypes from the whole labeled images $d^L$, for each domain and each class.\strut}
    \State \parbox[t]{210pt}{$\mathcal{S}_c^d = \{f(\mathbf{x}_i^d)|({y}_i=c)\}_{i=1}^{\lvert d^L \rvert}$ \Comment{Set of image features from the domain $d$ and class $c$}\strut}
    \State $\mathbf{K}_c^d = \frac{1}{\lvert \mathcal{S}_c^d  \rvert}\sum_{i=1}^{\lvert \mathcal{S}_c^d  \rvert} \mathcal{S}_c^d[i]$ 
    \State \# Compute the supervised loss
    \State $\mathcal{L}_s= CE(\mathcal{F}(\alpha (\mathbf{x_b}),y_b ))$
    \State \# Concat the labeled images to unlabeled (without labels)
    \State $u^{d}=[\textbf{u}^{d},\textbf{x}^{d}] \, \forall$ $d_i \in d$ 
    \For{ $u^{d_i}\in u^{d} $} 
        \State \# Generate pseudo label
        \If {$\max(\sigma(\mathcal{F}(u^{d_i}))) \ge \tau$}
        \State $\tilde{y} = \text{argmax}(\sigma(\mathcal{F}(u^{d_i})))$ 
        \State \# Compute unsupervised loss
         \State $\mathcal{L}_u=CE(\sigma(\mathcal{F}(u^{d_i})),\tilde{y})$
         \State \# Compute FBC loss
        \State $\mathbf{z}^{d_{i}} = \langle f(u^{d_i}), \mathbf{K}_{c}^{d_{i}} \rangle$ ; ${p}^{d_{i}} =\sigma(\mathbf{z}_i^{d_{i}})$
        \State $\mathbf{z}^{d_{j}} = \langle  f(u^{d_i}), \mathbf{K}_{c}^{d_{j}} \rangle$ ; ${p}^{d_{j}} =\sigma(\mathbf{z}^{d_{j}})$ 
        \State  $\mathcal{L}_{FBC}= CE({p}^{d_{i}},\tilde{y}) + CE({p}^{d_{j}},\tilde{y})$ \Comment{(Eq:~\ref{eq:feature_view_constraint})}
        \State \# Compute SA loss
        \State  $\phi^{d_i}=max(\mathbf{z}^{d_i})$
        \State $\Phi^{d_i}=\frac{1}{N-1}\sum_{n=2}^{N}\text{sort}(\mathbf{z}^{d_i})[n]$
        \State $\phi^{d_j}=\mathbf{z}^{d_j}[\text{argmax}(\mathbf{z}^{d_i})]$ 
        \State $\mathcal{L}_{SA}= (1 - \phi^{d_{i}} + \Phi^{d_i}) + 1 - \phi^{d_{j}}$ \Comment{(Eq:~\ref{eq:SA})}
    \EndIf  
        \State return $\mathcal{L}= \mathcal{L}_s +\mathcal{L}_u + \mathcal{L}_{FBC}+ \mathcal{L}_{SA}$
    \EndFor
\EndFor
\end{algorithmic}

\end{algorithm}
\begin{table*}[!ht]
\centering
\setlength{\tabcolsep}{15pt}
\scalebox{0.8}{
\begin{tabular}{lccccc}
\toprule
Model               & PACS  & OH & VLCS & DigitsDG & TerraInc. \\
\midrule
ERM                 & $51.2\pm3.0$  & $51.7\pm0.6$ & $67.2\pm1.8$      & $22.7\pm1.0$      & $22.9\pm3.0$   \\ \midrule
EntMin              & $55.9\pm4.1$  & $52.7\pm0.5$ & $66.5\pm1.0$       & $28.7\pm1.3$      & $21.4\pm3.5$        \\
MeanTeacher         & $53.3\pm4.0$  & $50.9\pm0.7$ & $66.4\pm1.0$      & $28.5\pm1.4$     & $20.9\pm2.9$         \\
FlexMatch           & $65.1\pm2.5$  & $48.8\pm0.3$ & $56.0\pm2.8$       & $59.0\pm2.0$      & $24.9\pm4.3$        \\
FreeMatch           & $72.8\pm1.2$  & $53.8\pm0.7$ & $60.3\pm1.7$       & $58.9\pm1.4$      & $23.5\pm2.7$      \\
FixMatch            & $73.4\pm1.3$  & $55.1\pm0.5$ & $69.9\pm0.6$       & $56.0\pm2.2$      & $28.9\pm2.3$        \\
StyleMatch          & $78.4\pm1.1$  & $56.3\pm0.3$ & $72.5\pm1.5$       & $55.7\pm1.6$      & $28.7\pm2.7$        \\  \midrule
FlexMatch + Ours    & $71.0\pm1.4$  & $51.3\pm0.1$ & $58.0\pm2.1$       & $\textbf{66.2}\pm\textbf{0.6}$      & $28.8\pm2.6$          \\
FreeMatch + Ours    & $73.7\pm3.6$  & $55.0\pm0.2$ & $62.1\pm1.4$       & $65.0\pm1.5$     & $26.5\pm3.2$         \\
FixMatch + Ours     & $77.3\pm1.1$  & $55.8 \pm0.2$ & $71.3\pm0.7$       & $62.0\pm1.5$   & $\textbf{33.2}\pm\textbf{2.0}$        \\
StyleMatch + Ours & $\textbf{79.3}\pm\textbf{0.9}$ & $\textbf{56.5}\pm\textbf{0.2}$ & $\textbf{72.9}\pm\textbf{0.7}$ & $58.7\pm1.7$  & $30.4\pm3.7$   \\
\bottomrule
\end{tabular}}
\vspace{-0.6em}
\caption{SSDG accuracy (\%) with 5 labels per class. (Average over 5 independent seeds is reported.) }
\vspace{-1em}
\label{tab:5_labels}
\end{table*}

\section{Experiments}
\label{section:Experiments}

\noindent\textbf{Datasets, training and implementation details:}
We utilize PACS \cite{li2017deeper}, OfficeHome \cite{venkateswara2017deep}, Digits \cite{zhou2020deep}, TerraIncognita \cite{beery2018recognition} and VLCS \cite{torralba2011unbiased} datasets which are widely used to report DG performance. For a detailed description of the datasets, refer to the supplementary materials. We conduct experiments under two settings; 10 labels and 5 labels per class while labeled images are selected randomly. The latter setting is more challenging due to the extreme scarcity of labeled data. 
Following \cite{zhou2023semi}, we randomly sample 16 labeled and 16 unlabeled images from each source domain to construct a minibatch. The labeled subset of minibatch is used to calculate the supervised loss while the (complete) minibatch, including both the labeled images (with ground truth labels dropped) and unlabeled images, are utilized to calculate all the unsupervised losses \cite{zhou2023semi}. We use ImageNet\cite{deng2009imagenet} pre-trained ResNet-18\cite{he2016deep} as the backbone architecture and a single-layer MLP head as the classifier. We use SGD as the optimizer with an initial learning rate of 0.003 for the backbone and 0.01 for the classifier, respectively. Both learning rates are decayed using the cosine annealing rule and we train all the models for 20 epochs on all datasets except TerraIncognita (trained for 10 epochs). We set $n=\lceil \frac{|C|}{2} \rceil$, from sec.~\ref{subsec: sematic_alignment_constraint}, in all experiments. 

\noindent\textbf{Evaluation protocol:}
\label{evaluation_protocol}
We use the leave-one-domain out protocol for evaluation which has been used widely in domain generalization \cite{Gulrajani2021InSO}. In this protocol, one domain is used as the target while the remaining domains are used as the source data to train the model. The target domain is unseen during the training phase and the model is evaluated on this unseen target domain. We report top-1 accuracy averaged over 5 independent trials.

\noindent\textbf{Baselines:}
Since our method combines SSL with DG, we select state-of-the-art methods in both paradigms. We choose EntMin \cite{grandvalet2004semi}, MeanTeacher \cite{tarvainen2017mean}, FixMatch \cite{sohn2020fixmatch}, FlexMatch \cite{zhang2021flexmatch} and  FreeMatch \cite{wang2023freematch} methods in SSL. StyleMatch\cite{zhou2023semi} is selected as an SSDG baseline as it shows promising performance under 5 and 10 labels settings. Further, we chose ERM\cite{vapnik1999nature} in DG as it shows competitive performance against many existing DG methods \cite{Gulrajani2021InSO}. For a detailed comparison with existing DG methods, see supplementary materials.

\subsection{Results}
\label{subsection:Results}

We report the performance of our method when integrated into four different baselines: StyleMatch, FixMatch, FlexMatch, and FreeMatch on five challenging DG datasets (Table~\ref{tab:10_labels},\ref{tab:5_labels}). \textbf{PACS:} Our method consistently improves the performance of all four baselines in both the 10 labels and 5 labels settings. We obtain 1.6\%, and 3.9\% average gains with our approach on top of the FixMatch baseline for 10 labels, and 5 labels setting respectively. Further, we attain 80.5\% average accuracy on 10 class settings, when applied with StyleMatch baseline. \noindent\textbf{VLCS:} Our method is capable of delivering gain over all four baselines in both the 10 labels and 5 labels per class settings. Specifically, when combined with FixMatch baseline our method achieves more than 1.8\% average gain on both settings. \noindent\textbf{OfficeHome:} Our approach improves the FixMatch baseline by 1.2\% on average in 10 labels settings. In the same setting, our method, in tandem with StyleMatch, achieves the best figures across all domains with an average of 60.3\%. \noindent\textbf{Digits-DG:} When combined with FixMatch baseline, our approach improves FixMatch by a significant gain of 4.0\% in 10 label setting and 6.0\% in the 5-label setting. On average, we consistently improve all the baselines by approximately 5\%  with our approach. \noindent\textbf{Terra Incognita:} Our method provides gain over all four baselines on the Terra Incognita dataset which has a number of real-world distribution shifts such as illumination shifts, blur, shifts in the size of the region of interest (ROI), occlusions, camouflage and shifts in perspective. In both 5 and 10 labels settings, we obtain over 4.0\% gain with our approach over the FixMatch baseline. \\
Note that, StyleMatch is unable to improve on its baseline FixMatch when there are no style distributions present in the source domains (on datasets such as Digits and TerraIncognita) where our proposed method demonstrates significant gains with over 4.0\% improvement. As our approach is model agnostic, it can be seamlessly integrated with different SSL and SSDG baselines. Further, the improvements over the baselines are consistent across all the DG datasets under both 10 and 5 labels per class settings. 

\begin{table}[!tp]
\centering
\scalebox{0.8}{
\begin{tabular}{lc}
\toprule
Method                                                                                      & Average \\ \midrule
Baseline \cite{sohn2020fixmatch}                                                                         & 73.4     \\
Baseline + $\mathcal{L}_{\mathrm{FBC(same-domain)}}$                                        & 76.0       \\
Baseline + $\mathcal{L}_{\mathrm{FBC(different-domain)}}$                                   & 74.9         \\
Baseline + $\mathcal{L}_{\mathrm{FBC}}$                                                     &  76.7       \\
Baseline + $\mathcal{L}_{\mathrm{SA}}$                                                      &  74.8       \\
Baseline + $\mathcal{L}_{\mathrm{FBC}}$ + $\mathcal{L}_{\mathrm{SA(same-domain)}}$          & 77.0           \\
Baseline + $\mathcal{L}_{\mathrm{FBC}}$ + $\mathcal{L}_{\mathrm{SA}}$ (Ours)               & \textbf{77.3}          \\     \bottomrule   

\end{tabular}}
\caption{Ablation study on PACS (5 labels per class).}
\label{tab:component-wise_ablation}
\vspace*{-\baselineskip}
\end{table}

\begin{table*}[!htp]
\centering
\small
\scalebox{0.75}{
\begin{tabular}{lccccccccc}
\toprule
\multicolumn{1}{c}{\multirow{2}{*}{Domain Shift}} & 
\multicolumn{1}{c}{\multirow{2}{*}{Dataset}} &
\multicolumn{8}{c}{Method} \\
\cmidrule{3-10} \\ [-1em] & & 
FlexMatch & FreeMatch & FixMatch & StyleMatch & FlexMatch+Ours & FreeMatch+Ours & FixMatch+Ours & StyleMatch+Ours 
 \\ \midrule
Style Shifts                  & OH, PACS                   & 56.9                       & 63.3                      & 64.1                     & 64.25                      & 61.2                            & 64.4                             & 66.6                             & \textbf{67.9 }                             \\
Background Shifts             & VLCS, Digits               & 57.6                       & 59.6                      & 62.9                     & 64.1                      & 62.1                            & 63.6                             & \textbf{66.7}                            & 65.8                                 \\
Corruption Shift              & Terra                      & 24.9                       & 23.5                      & 28.9  & 28.7                      & 28.8                            & 26.5                             & \textbf{33.2 }                            & 30.4                              \\
\bottomrule
\end{tabular}}
\caption{Accuracy(\%) for different types of domain shifts in 5 labels per class setting. Corruption shifts include changes in illumination, changes in perspective, changes in ROI size, camouflage, occlusions, and blur which are often present in real-world distributions.}
\label{tab:domain-shifts}
\vspace*{-\baselineskip}
\end{table*}

\subsection{Ablation Study and Analysis}
\label{subsection:Ablation Study and Analysis}

\noindent \textbf{Impact of different components:} In
Table~\ref{tab:component-wise_ablation} we report the performance contribution of individual components in our approach. We note the following important trends: (1) each component is capable of improving the performance over the baseline, (2) the feature-based conformity component, which is the first component of our proposed method, provides a high gain of 3.3\% while showing each same domain and different domain feature alignment is capable of improving the baseline's performances individually. (3) our proposed method provides the best gain of 3.9\% when the feature-based conformity is coupled with the semantic alignment constraint. 

\noindent \textbf{Performance under various domain shifts:}
We evaluate the performance under various domain shifts (Table ~\ref{tab:domain-shifts}) including background shifts, style shifts, and corruption shifts which occur in real-world scenarios. Our approach provides visible gains over baselines under various domain shifts. Existing SSDG methods such as StyleMatch operate within the confines of the input pixel space and under the assumption that some style variations are present between source domains. Unlike our method which harnesses the information at feature space, such methods fail to generalize when they encounter real-world corruption shifts such as shifts in illumination, perspective, ROI size, blur, camouflage, and occlusions in the source domains.

\noindent \textbf{Feature visualization:} We visualize the feature space for PACS test domains with FixMatch as baseline (see Fig.~\ref{fig:tsne} (left)). When FixMatch is coupled with our first component, feature-based conformity, we observe better class-wise discrimination over the baseline. When the feature-based conformity is supported by the semantic alignment constraint, we can observe well-separated and well-compact class-wise clusters in feature space that help improve the classification performance. Also, our proposed method encourages classes to be orthogonal in the feature space (Fig.~\ref{fig:tsne}(right)).

\section{Conclusion and Limitations}
\vspace{-0.5pt}
We approach the relatively unexplored problem of semi-supervised domain generalization (SSDG) and propose a new method, built with feature-based conformity and semantics alignment constraint modules, towards addressing the key challenges in SSDG. The feature-based conformity mechanism aligns the posterior distributions from two views, while the semantics alignment constraint further boosts the effectiveness of feature-based conformity by regularizing the semantic layout of feature space. Our approach is plug-and-play, parameter-free and model-agnostic, so it can be seamlessly integrated into different baselines as validated in our results. Extensive experiments on different challenging DG benchmarks show that our method delivers a consistent and notable gain over four recent baselines.
An aspect to consider is that our approach may necessitate significant adjustments to accommodate semi-supervised single-source DG.

{ \small
\bibliographystyle{ieeenat_fullname}
    \bibliography{main}
}



\end{document}